\documentclass{article}
\usepackage{ijcai25}

\usepackage{times}
\usepackage{soul}
\usepackage{url}
\usepackage[hidelinks]{hyperref}
\usepackage[utf8]{inputenc}
\usepackage[small]{caption}
\usepackage{graphicx}
\usepackage{amsmath}
\usepackage{amsthm}
\usepackage{booktabs}
\usepackage{algorithm}
\usepackage{algorithmic}
\usepackage[switch]{lineno}

\usepackage{booktabs}						
\usepackage{multirow}						
\usepackage{amsfonts}						
\usepackage{graphicx}						
\usepackage{duckuments}						

\usepackage{wrapfig}

\usepackage{multirow}
\usepackage{clipboard}

\usepackage{enumitem}

\title{What Do LLMs Need to Understand Graphs: \\A Survey of Parametric Representation of Graphs}

\author{
Dongqi Fu\thanks{Equal Contribution}\and
Liri Fang$^*$\and
Zihao Li$^*$\and
Hanghang Tong\and
Vetle I. Torvik\and
Jingrui He
\affiliations
University of Illinois Urbana-Champaign\\
\emails
\{dongqif2, lirif2, zihaoli5, htong, vtorvik, jingrui\}@illinois.edu
}

\begin{document}

\maketitle

\begin{abstract}
Graphs, as a relational data structure, have been widely used for various application scenarios, like molecule design and recommender systems. 
Recently, large language models (LLMs) are reorganizing in the AI community for their expected reasoning and inference abilities.
Making LLMs understand graph-based relational data has great potential, including but not limited to (1) distillate external knowledge base for eliminating hallucination and breaking the context window limit for LLMs' inference during the retrieval augmentation generation process; (2) taking graph data as the input and directly solve the graph-based research tasks like protein design and drug discovery.
However, \textbf{inputting the entire graph data to LLMs is not practical} due to its complex topological structure, data size, and the lack of effective and efficient semantic graph representations.
A natural question arises: \textit{Is there a kind of graph representation that can be described by natural language for LLM's understanding and is also easy to require to serve as the raw input for LLMs?}
Based on statistical computation, \textbf{graph laws} pre-defines a set of parameters (e.g., degree, time, diameter) and identifies their relationships and values by observing the topological distribution of plenty of real-world graph data. We believe this kind of parametric representation of graphs, graph laws, can be a solution for making LLMs understand graph data as the input.
In this survey, we first review the previous study of graph laws from multiple perspectives, i.e., \textit{macroscope and microscope} of graphs, \textit{low-order and high-order} graphs, \textit{static and dynamic} graphs, \textit{different observation spaces}, and \textit{newly proposed graph parameters}. After we review various real-world applications benefiting from the guidance of graph laws, we conclude the paper with current challenges and future research directions.
\end{abstract}

\section{Introduction}
\subsection{Motivation of this Paper}
\begin{figure}[h!]
    \centering
    \includegraphics[width=0.5\textwidth]{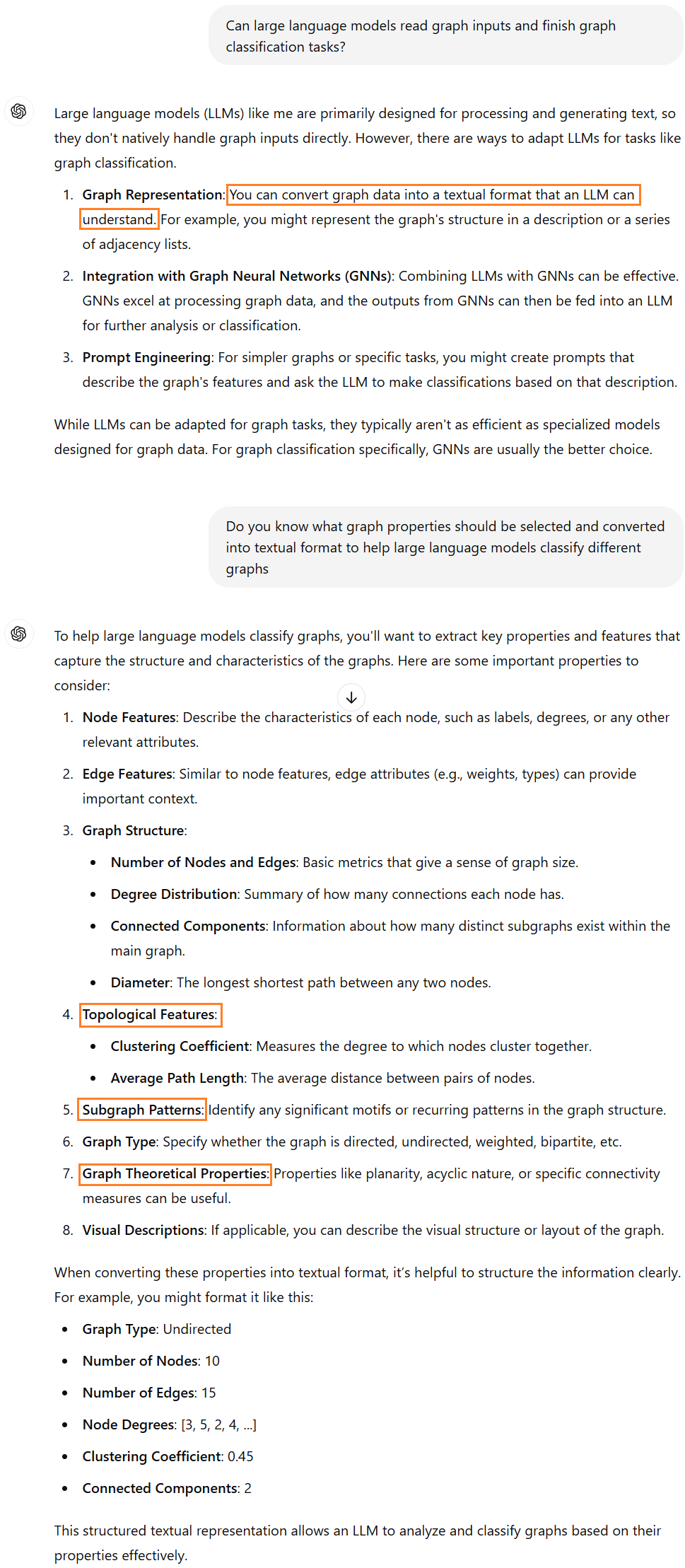}
    \vspace{-5mm}
    \caption{A Case Study of ChatGPT 4o's Explaining about Why it Needs Graph Laws}
    \vspace{-5mm}
    \label{fig:motivation}
\end{figure}

Graphs serve as a fundamental relational data structure and are extensively utilized in a wide range of application scenarios, including molecule design, social network analysis, and recommender systems~\cite{zhou2020graph}. Their ability to represent complex interconnections among entities makes them indispensable in modeling real-world relationships. However, despite their widespread use, integrating graph-based data input with large language models (LLMs) remains a challenging problem, as shown in Figure~\ref{fig:motivation} with a ChatGPT 4o~\cite{hurst2024gpt} case study.

Recently, LLMs have revolutionized the AI community with their remarkable reasoning and inference capabilities~\cite{touvron2023llama,guo2025deepseek}, and those models have demonstrated significant potential in various tasks, such as natural language understanding, machine translation, and knowledge extraction. Given the growing importance of LLMs, enabling them to comprehend and process graph-based relational data could open new frontiers in artificial intelligence research and applications. This integration holds immense potential for enhancing LLMs in multiple ways, including but not limited to:
\begin{itemize}[leftmargin=*]
    \item \textbf{Knowledge Distillation for LLMs}: Graph-based external knowledge bases can provide crucial insights, mitigating issues such as hallucinations in LLM-generated responses and overcoming the limitations imposed by fixed context windows. By incorporating structured graph data, LLMs can improve retrieval-augmented generation (RAG) techniques and enhance inference accuracy~\cite{edge2024local,he2024g}.
    \item \textbf{Direct Graph-Based Problem Solving}: Many research domains, such as protein design and drug discovery, inherently rely on graph-based data representations~\cite{liang2023drugchat,wang2024graph2token}. Equipping LLMs with the capability to understand and manipulate graph structures could significantly advance research in these fields by enabling direct problem-solving approaches.
\end{itemize}

Despite the clear advantages of incorporating graph data into LLMs, several challenges hinder this integration. The primary obstacles include (1) the complexity of graph topologies, (2) the size of graph datasets, and (3) the absence of effective semantic representations of graphs that LLMs can process efficiently. Unlike textual data, which LLMs are inherently designed to understand, graphs lack a straightforward natural language representation. This leads to a fundamental research question: \textit{Is there a form of graph representation that is both interpretable in natural language for LLMs and compact enough to serve as a viable input format?}

A promising solution lies in the concept of \textbf{graph laws}. Graph laws refer to statistical principles that define relationships between key structural parameters of graphs, such as degree, clustering coefficients, diameter, and time. \textbf{Hence, a graph can be represented by a few parameters to reflect its properties well}.
Correspondingly, the parameters' formal mathematical relations and values are captured by graph law researchers by analyzing real-world, large-scale graph data distributions. By encoding graph properties through predefined sets of parameters, graph laws offer a way to translate complex graph topologies into a form that LLMs can potentially comprehend.
For example, according to the previous research~\cite{DBLP:conf/kdd/LeskovecKF05,DBLP:conf/kdd/LeskovecBKT08}, the relation between the possibility of a newly-arrived node connecting to an old node (parameter \#1) and the degree of that old node (parameter \#2) is studied by maximum likelihood estimation (MLE) based on the observed real-world graph data~\cite{DBLP:conf/kdd/LeskovecBKT08}.

\subsection{Background of Graph Representations}
\begin{figure}[h]
    \centering
    \includegraphics[width=0.46\textwidth]{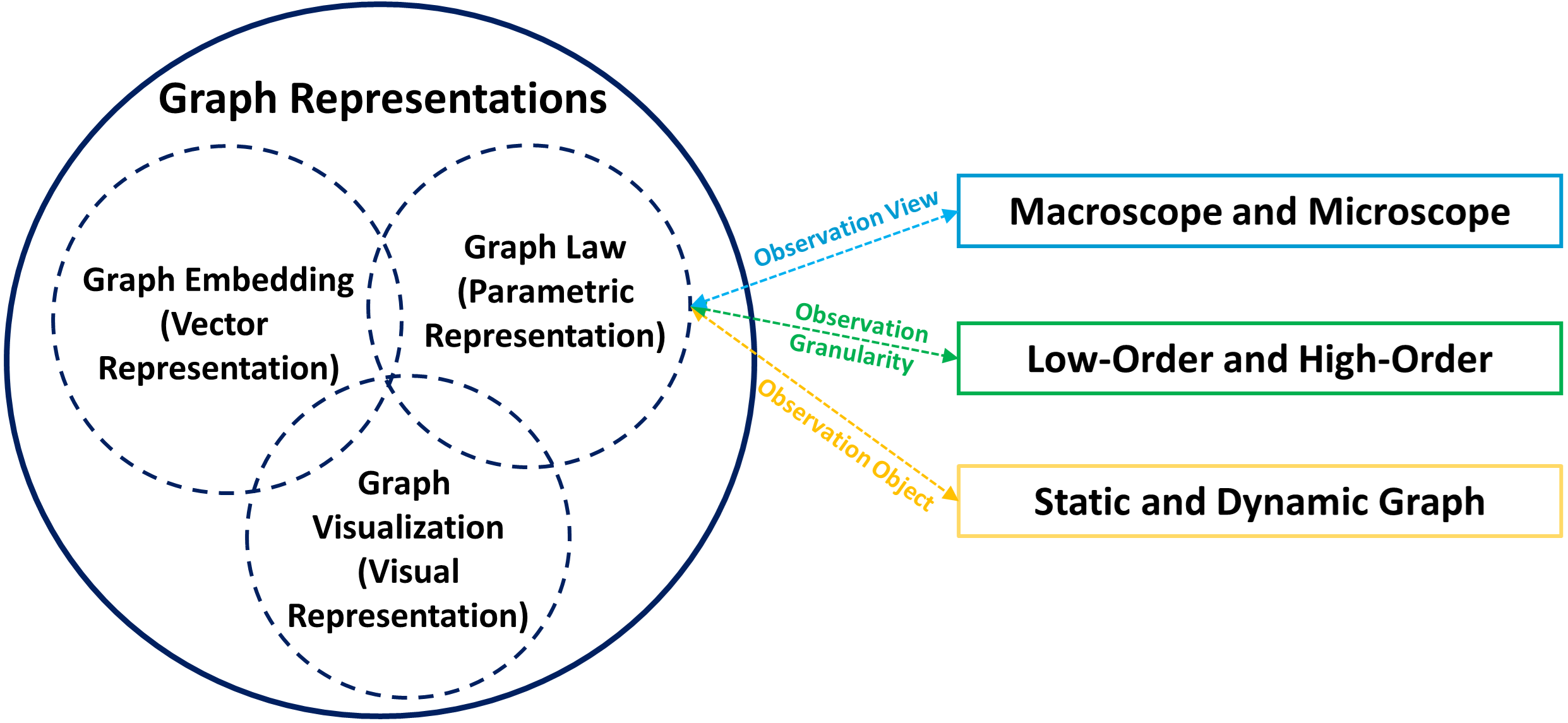}
    \caption{Position of Graph Law in Graph Representations.}
    \label{fig:introduction}
\end{figure}

To model real-world tasks within graphs, graph representations are indispensable middleware that provides the basis for specific and complex task-oriented computations.
To be specific, graph representations can be decomposed into three aspects as shown in Figure~\ref{fig:introduction}, (1) graph embedding (i.e., vector representation), (2) graph law (i.e., parametric representation), and (3) graph visualization (i.e., visual representation).

First of all, graph representations can be in the form of embedding matrices, i.e., the graph topological information and attributes are encoded into matrices, which has been widely discussed and studied in the research community and usually can be referred to as \textit{graph representation learning}.~\cite{hamilton2020graph}.
Then, graphs can also be represented by plotting directly for a better human-understandable illustration. For example, one interesting research topic is how to plot the graph topological structures into the 2D space with less structure distortion. More interesting works can be referred to~\cite{fu2022natural}.
Last but not least, graphs can also be represented by a few key parameters such as Erd\H{o}s-R\'enyi random graph $G(n,p)$~\cite{DBLP:journals/csur/DrobyshevskiyT20}, where $n$ stands for the number of nodes in the graph, and $p$ stands for the independent edge connection probability in the graph.
As shown in Figure~\ref{fig:introduction}, these three representations can have overlapping to mutually contribute to each other~\cite{fu2022natural}.

\subsection{Organization of this Paper}
Graph law is the study of investigating the statistical properties of graphs.
In this survey, we introduce the graph laws studies in the macroscopic view and microscopic view, plus multiple angles like low-order and high-order connections, static and dynamic graphs, as shown in Table~\ref{tab:summarization}.
\begin{itemize}[leftmargin=*]
    \item \textbf{Macroscopic and Microscopic Views}. The macroscopic graph laws describe the graph properties in a global view, like how the total degree (or eigenvalues) distribution of the entire graph looks like~\cite{DBLP:conf/kdd/LeskovecKF05}; while the microscopic laws try to focus on the individual behavior and investigate their behaviors as part in the entire graph~\cite{DBLP:conf/kdd/LeskovecBKT08}.
    
    \item \textbf{Low-Order and High-Order Connections}. Most graph laws are based on the node-level connections (i.e., low-order connections), while some graph law investigations are based on the group activities (high-order connections), i.e., motif in~\cite{DBLP:conf/wsdm/ParanjapeBL17,DBLP:conf/www/ZenoFN20}, hyperedge in~\cite{DBLP:conf/kdd/DoYHS20,DBLP:conf/icdm/KookKS20}, and simplex in~\cite{DBLP:journals/pnas/BensonASJK18,comrie2021hypergraph}.
    
    \item \textbf{Static and Dynamic Graphs}. Compared with static graphs, dynamic graphs allow the graph components like topology structure and node attributes to evolve over time. Correspondingly, some graph laws study how the graph parameters change over time and their temporal relations. Note that, in some graph research, the dynamics are created by the algorithms, like adding virtual nodes to preserve graph representations~\cite{abraham2007reconstructing,DBLP:conf/icml/LiuCSJ22}; this kind of research is beyond the discussion of this paper, and we focus on the graph law with natural time.
\end{itemize}

Furthermore, in Section~\ref{sec:applications}, we survey different real-world applications that would benefit from the guidance of graph laws. 
In Section~\ref{sec: new params}, we introduce different observation spaces and newly proposed parameters, whose corresponding laws are not fully discovered yet. 
Also, in Section~\ref{sec:related_work}, we review the related work of graph law surveys and compare the differences.
Finally, in Section~\ref{sec:future_directions}, we conclude the survey with current challenges and future directions.

\renewcommand\arraystretch{2}

\begin{table*}[t]
\caption{A summary of parameteric representations of graphs. Some laws have multiple aspects and are indexed by numbers in parentheses.}
\vspace{-2mm}
\label{tab:summarization}
\begin{center}
\scalebox{0.58}{
\begin{tabular}{lllllll}
\toprule
\multicolumn{1}{l|}{\textbf{Input}}                      & \multicolumn{1}{l|}{\textbf{Law}}  & \multicolumn{1}{l|}{\textbf{Parameter}} & \multicolumn{1}{l|}{\textbf{Scope}} & \multicolumn{1}{l|}{\textbf{Order}} & \multicolumn{1}{l|}{\textbf{\small{Temporality}}} & \multicolumn{1}{l}{\textbf{Description}} \\ \midrule
\multicolumn{1}{l|}{\multirow{16}{*}{Graphs}} & \multicolumn{1}{l|}{Densification Law~\cite{DBLP:conf/kdd/LeskovecKF05}} &  \multicolumn{1}{l|}{Density degree $\alpha$} & \multicolumn{1}{l|}{Macro} & \multicolumn{1}{l|}{Low} & \multicolumn{1}{l|}{Dynamic} & \multicolumn{1}{l}{$e(t) \propto n(t)^{\alpha}, \alpha \in [1, 2]$, $e(t)$ is \# edges at $t$} \\ \cline{2-7}
\multicolumn{1}{l|}{} & \multicolumn{1}{l|}{Shrinking Law~\cite{DBLP:conf/kdd/LeskovecKF05}} & \multicolumn{1}{l|}{Effective diameter $d$} & \multicolumn{1}{l|}{Macro} &\multicolumn{1}{l|}{Low} & \multicolumn{1}{l|}{Dynamic} & \multicolumn{1}{l}{$d_{t+1} < d_t$, $d$ decreases as network grows} \\ \cline{2-7}
\multicolumn{1}{l|}{} & \multicolumn{1}{l|}{Motif Differing Law(1)~\cite{DBLP:conf/wsdm/ParanjapeBL17}} & \multicolumn{1}{l|}{Numbers of similar motifs $n$} &  \multicolumn{1}{l|}{Macro} &\multicolumn{1}{l|}{High} & \multicolumn{1}{l|}{Dynamic} & \multicolumn{1}{l}{$n_1 \neq n_2$ for different domains} \\ \cline{2-7}
\multicolumn{1}{l|}{} & \multicolumn{1}{l|}{Motif Differing Law(2)~\cite{DBLP:conf/wsdm/ParanjapeBL17}} & \multicolumn{1}{l|}{Motif occurring timestamp $t$} &  \multicolumn{1}{l|}{Macro} &\multicolumn{1}{l|}{High} & \multicolumn{1}{l|}{Dynamic} & \multicolumn{1}{l}{$t_1 \neq t_2$ for different motifs} \\ \cline{2-7}
\multicolumn{1}{l|}{} & \multicolumn{1}{l|}{Egonet Differing Law~\cite{DBLP:journals/pnas/BensonASJK18}} & \multicolumn{1}{l|}{Features of Egonets $X$} &  \multicolumn{1}{l|}{Macro} &\multicolumn{1}{l|}{High} & \multicolumn{1}{l|}{Static} & \multicolumn{1}{l}{$X_1 \neq X_2$ for different domains} \\ \cline{2-7}
\multicolumn{1}{l|}{} & \multicolumn{1}{l|}{Simplicial Closure Law~\cite{DBLP:journals/pnas/BensonASJK18}} & \multicolumn{1}{l|}{Simplicial closure probability $p$} &  \multicolumn{1}{l|}{Macro} &\multicolumn{1}{l|}{High} & \multicolumn{1}{l|}{Static} & \multicolumn{1}{l}{$p$ increases with additional edges or tie strength} \\ \cline{2-7}
\multicolumn{1}{l|}{} & \multicolumn{1}{l|}{Spectral Power Law(1)~\cite{DBLP:conf/kdd/EikmeierG17}} & \multicolumn{1}{l|}{Degree, SVD, eigen distributions} &  \multicolumn{1}{l|}{Macro} &\multicolumn{1}{l|}{High} & \multicolumn{1}{l|}{Static} & \multicolumn{1}{l}{These distributions usually follow power-law} \\ \cline{2-7}
\multicolumn{1}{l|}{} & \multicolumn{1}{l|}{Spectral Power Law(2)~\cite{DBLP:conf/kdd/EikmeierG17}} & \multicolumn{1}{l|}{Degree, SVD, eigen distributions} &  \multicolumn{1}{l|}{Macro} &\multicolumn{1}{l|}{High} & \multicolumn{1}{l|}{Static} & \multicolumn{1}{l}{If one follow power-law, usually others follow} \\ \cline{2-7}
\multicolumn{1}{l|}{} & \multicolumn{1}{l|}{Edge Attachment Law(1)~\cite{DBLP:conf/kdd/LeskovecBKT08}} & \multicolumn{1}{l|}{Node degree $d$, edge create $p_e(d)$} &  \multicolumn{1}{l|}{Micro} &\multicolumn{1}{l|}{Low} & \multicolumn{1}{l|}{Dynamic} & \multicolumn{1}{l}{$p_e(d) \propto d$ for node with degree $d$} \\ \cline{2-7}
\multicolumn{1}{l|}{} & \multicolumn{1}{l|}{Edge Attachment Law(2)~\cite{DBLP:conf/kdd/LeskovecBKT08}} & \multicolumn{1}{l|}{Node age $a(u)$, edge create $p_e(d)$} &  \multicolumn{1}{l|}{Micro} &\multicolumn{1}{l|}{Low} & \multicolumn{1}{l|}{Dynamic} & \multicolumn{1}{l}{$p_e(d) $ seems to be non-decreasing with $a(u)$} \\ \cline{2-7}
\multicolumn{1}{l|}{} & \multicolumn{1}{l|}{Triangle Closure Law(1)~\cite{DBLP:journals/tkdd/HuangDTYCF18}} & \multicolumn{1}{l|}{Triangular connections $e_1, e_2, e_3$} &  \multicolumn{1}{l|}{Micro} &\multicolumn{1}{l|}{Low} & \multicolumn{1}{l|}{Dynamic} & \multicolumn{1}{l}{Strong $e_3 \Rightarrow$ unlikely $e_1/e_2$ will be weakened} \\ \cline{2-7}
\multicolumn{1}{l|}{} & \multicolumn{1}{l|}{Triangle Closure Law(2)~\cite{DBLP:journals/tkdd/HuangDTYCF18}} & \multicolumn{1}{l|}{Triangular connections $e_1, e_2, e_3$} &  \multicolumn{1}{l|}{Micro} &\multicolumn{1}{l|}{Low} & \multicolumn{1}{l|}{Dynamic} & \multicolumn{1}{l}{Strong $e_1/e_2 \Rightarrow$ unlikely they will be weakened} \\ \cline{2-7}
\multicolumn{1}{l|}{} & \multicolumn{1}{l|}{Local Closure Law~\cite{DBLP:conf/wsdm/YinBL19}} & \multicolumn{1}{l|}{Local closure coefficient $H(u)$} &  \multicolumn{1}{l|}{Micro} &\multicolumn{1}{l|}{Low} & \multicolumn{1}{l|}{Static} & \multicolumn{1}{l}{Please refer to Section \ref{sec: new params} for details} \\ \cline{2-7}
\multicolumn{1}{l|}{} & \multicolumn{1}{l|}{Spectral Density Law~\cite{DBLP:conf/kdd/DongBB19}} & \multicolumn{1}{l|}{Density of states $\mu(\lambda)$} &  \multicolumn{1}{l|}{Macro} &\multicolumn{1}{l|}{High} & \multicolumn{1}{l|}{Static} & \multicolumn{1}{l}{Please refer to Section \ref{sec: new params} for details} \\ \cline{2-7}
\multicolumn{1}{l|}{} & \multicolumn{1}{l|}{Motif Activity Law(1)~\cite{DBLP:conf/www/ZenoFN20}} & \multicolumn{1}{l|}{Motif type} &  \multicolumn{1}{l|}{Micro} &\multicolumn{1}{l|}{High} & \multicolumn{1}{l|}{Dynamic} & \multicolumn{1}{l}{Motifs do not transit from one type to another} \\ \cline{2-7}
\multicolumn{1}{l|}{} & \multicolumn{1}{l|}{Motif Activity Law(2)~\cite{DBLP:conf/www/ZenoFN20}} & \multicolumn{1}{l|}{Motif re-appear rate} &  \multicolumn{1}{l|}{Micro} &\multicolumn{1}{l|}{High} & \multicolumn{1}{l|}{Dynamic} & \multicolumn{1}{l}{Motifs re-appear with configured rates} \\ \midrule
\multicolumn{1}{l|}{\multirow{6}{*}{\small{Hypergraphs}}} & \multicolumn{1}{l|}{Degree Distribution Law~\cite{DBLP:conf/kdd/DoYHS20}} &  \multicolumn{1}{l|}{Node degree, edge link probability} &\multicolumn{1}{l|}{Macro} & \multicolumn{1}{l|}{High} & \multicolumn{1}{l|}{Dynamic} & \multicolumn{1}{l}{High-degree nodes are likely to form new links}            \\ \cline{2-7}
\multicolumn{1}{l|}{} & \multicolumn{1}{l|}{SVD Distribution Law~\cite{DBLP:conf/kdd/DoYHS20}} & \multicolumn{1}{l|}{Singular value distribution} & \multicolumn{1}{l|}{Macro} & \multicolumn{1}{l|}{High} & \multicolumn{1}{l|}{Static} & \multicolumn{1}{l}{Singular value distribution usually heavy-tailed}            \\ \cline{2-7}
\multicolumn{1}{l|}{} & \multicolumn{1}{l|}{Diminishing Overlaps~\cite{DBLP:conf/icdm/KookKS20}} & \multicolumn{1}{l|}{density of interactions $DoI(\mathcal{H}(t))$} &  \multicolumn{1}{l|}{Macro} &\multicolumn{1}{l|}{High} & \multicolumn{1}{l|}{Dynamic} & \multicolumn{1}{l}{Overall hyperedge overlaps decrease over time} \\ \cline{2-7}
\multicolumn{1}{l|}{} & \multicolumn{1}{l|}{Densification Law~\cite{DBLP:conf/icdm/KookKS20}} & \multicolumn{1}{l|}{Density degree $\alpha$} & \multicolumn{1}{l|}{Macro} &\multicolumn{1}{l|}{High} & \multicolumn{1}{l|}{Dynamic} & \multicolumn{1}{l}{$e(t) \propto n(t)^{\alpha}, \alpha \geq 1$, $e(t)$ is \# hyperedges at $t$}            \\ \cline{2-7}
\multicolumn{1}{l|}{} & \multicolumn{1}{l|}{Shrinking Law~\cite{DBLP:conf/icdm/KookKS20}} & \multicolumn{1}{l|}{Hypergraph effective diameter $d$} & \multicolumn{1}{l|}{Macro} &\multicolumn{1}{l|}{High} & \multicolumn{1}{l|}{Dynamic} & \multicolumn{1}{l}{$d_{t+1} < d_t$, $d$ decreases as network grows} \\ \cline{2-7}
\multicolumn{1}{l|}{} & \multicolumn{1}{l|}{Edge Interacting Law~\cite{comrie2021hypergraph}} & \multicolumn{1}{l|}{Edge interacting rate} &  \multicolumn{1}{l|}{Micro} &\multicolumn{1}{l|}{High} & \multicolumn{1}{l|}{Dynamic} & \multicolumn{1}{l}{Temporally adjacent interactions highly similar}            \\ \midrule
\multicolumn{1}{l|}{\multirow{2}{*}{\small{Heterographs}}} & \multicolumn{1}{l|}{Densification Law~\cite{DBLP:conf/aaai/WangZS19a}} &  \multicolumn{1}{l|}{Density degree $\alpha$, \# meta-path} &\multicolumn{1}{l|}{Macro} & \multicolumn{1}{l|}{Low} & \multicolumn{1}{l|}{Dynamic} & \multicolumn{1}{l}{$e(t) \propto n(t)^{\alpha}, \alpha \geq 1$ for some meta-path}            \\ \cline{2-7}
\multicolumn{1}{l|}{} & \multicolumn{1}{l|}{Non-densification Law~\cite{DBLP:conf/aaai/WangZS19a}} & \multicolumn{1}{l|}{Density degree $\alpha$, \# meta-path} & \multicolumn{1}{l|}{Macro} & \multicolumn{1}{l|}{Low} & \multicolumn{1}{l|}{Dynamic} & \multicolumn{1}{l}{Maybe, for some meta-path, $e(t) \not\propto n(t)^\alpha$}            \\ \midrule
\bottomrule
\end{tabular}
}
\end{center}
\vspace{-6mm}
\end{table*}

\renewcommand\arraystretch{2}

\section{Macroscopic Graph Laws}
In this section, we introduce the graph laws from the macroscope and microscope. In detail, we will introduce what is the intuition of researchers proposing or using graph statistical properties as parameters and how they fit the value of parameters against real-world observations.

Several classical theories model the growth of graphs, for example, Barabasi-Albert model~\cite{barabasi1999emergence,barabasi2002evolution} assumes that the graphs follow the uniform growth pattern in terms of the number of nodes, and the Bass model~\cite{mahajan1990new} and the Susceptible-Infected model~\cite{anderson1992infectious} follow the Sigmoid growth (more random graph models can be founded in~\cite{DBLP:journals/csur/DrobyshevskiyT20}). However, these pre-defined graph growths have been tested that they could not handle the complex real-world network growth patterns very well~\cite{DBLP:conf/www/KwakLPM10,DBLP:journals/tkde/ZangCFZ18}. To this end, researchers begin to fit the graph growth on real-world networks directly, to discover graph laws.

\subsection{Low-Order Macroscopic Laws} Based on fitting nine real-world temporal graphs from four different domains, the authors in~\cite{DBLP:conf/kdd/LeskovecKF05} found two temporal graph laws, called (1) \textit{Densification Laws} and (2) \textit{Shrinking Diameters}, respectively. First, the densification law states as follows.
\begin{equation}
    e(t) \propto n(t)^{\alpha}
\end{equation}
where $e(t)$ denotes the number of edges at time $t$, $n(t)$ denotes the number of nodes at time $t$, $\alpha \in [1,2]$ is an exponent representing the density degree. The second law, shrinking diameters, states that \textit{the effective diameter is decreasing as the network grows, in most cases}. Here, the diameter means the node-pair shortest distance, and the effective diameter of the graph means the minimum distance $d$ such that approximately 90\% of all connected pairs are reachable by a path of length at most $d$. Later, in~\cite{DBLP:journals/tkde/ZangCFZ18}, the densification law gets in-depth confirmed on four different real social networks, the research shows that the number of nodes and number of edges both grown exponentially with time, i.e., following the power-law distribution.  

\subsection{High-Order Macroscopic Laws} Above discoveries are based on the node-level connections (i.e., low-order connections), then several researchers start the investigation based on the group activities, for example, motifs~\cite{DBLP:conf/wsdm/ParanjapeBL17}, simplices~\cite{DBLP:journals/pnas/BensonASJK18} and hyperedges~\cite{DBLP:conf/kdd/DoYHS20,DBLP:conf/icdm/KookKS20}. Motif is defined as a subgraph induced by a sequence of selected temporal edges in~\cite{DBLP:conf/wsdm/ParanjapeBL17}, where the authors discovered that \textit{different domain networks have significantly different numbers of similar motifs, and different motifs usually occur at different time}. Similar laws are also discovered in~\cite{DBLP:journals/pnas/BensonASJK18} that the authors study 19 graph data sets from domains like biology, medicine, social networks, and the web, to characterize how high-order structure emerges and differs in different domains. They discovered that the higher-order Egonet features can discriminate the domain of the graph, and the probability of simplicial closure events typically increases with additional edges or tie strength. 

In hypergraphs, each hyperedge could connect an arbitrary number of nodes, rather than two~\cite{DBLP:conf/kdd/DoYHS20}, where the authors found that real-world static hypergraphs obey the following properties: (1) \textit{Giant Connected Components}, that there is a connected component comprising a large proportion of nodes, and this proportion is significantly larger than that of the second-largest connected component. (2) \textit{Heavy-Tailed Degree Distributions}, that high-degree
nodes are more likely to form new links. (3) \textit{Small Effective Diameters}, that most connected pairs can be reached by a small distance (4) \textit{High Clustering Coefficients}, that the global average of local clustering coefficient is high. (5) \textit{Skewed Singularvalue Distributions}, that the singular-value distribution is usually heavy-tailed. Later, the evolution of real-word hypergraphs is investigated in~\cite{DBLP:conf/icdm/KookKS20}, and the following laws are discovered.
\begin{itemize} [leftmargin=*]
    \item \textit{Diminishing Overlaps}: The overall overlaps of hyperedges decrease over time.
    \item \textit{Densification}: The average degrees increase over time.
    \item \textit{Shrinking Diameter}: The effective diameters decrease over time.
\end{itemize}
To be specific, given a hypergraph $G(t) = (V(t), E(t))$, the density of interactions is stated as follows.
\begin{equation}
    DoI(G(t)) = \frac{|~ \{\{e_i, e_j\} ~|~ e_{i} \cap e_{j} \neq \emptyset ~\text{for}~ e_{i}, e_{j} \in E(t)\}~|}{\{ \{ e_{i}, e_{j} \} | e_{i}, e_{j} \in E(t) \}}
\end{equation}
and the densification is stated as follows.
\begin{equation}
    |E(t)| \propto |V(t)| ^{s}
\end{equation}
where $s > 1$ stands for the density term.

In heterogeneous information networks (where nodes and edges can have multiple types), the power law distribution is also discovered~\cite{DBLP:conf/aaai/WangZS19a}. For example, for the triplet "author-paper-venue" (i.e., A-P-V), the number of authors is power law distributed w.r.t the number of A-P-V instances composed by an author.

\section{Microscopic Graph Laws}
In contrast to representing the whole distribution of the entire graph, many researchers try to model individual behavior and investigate how they interact with each other to see the evolution pattern microscopically.

\subsection{Low-Order Microscopic Laws} In~\cite{DBLP:conf/kdd/LeskovecBKT08}, the authors view temporal graphs in a three-fold process, i.e., node arrival (determining how many nodes will be added), edge initiation (how many edges will be added), and edge destination (where are each edge will be added). They ignore the deletion of nodes and edges, and they assign variables (models) to parameterize this process.
\begin{itemize}[leftmargin=*]
    \item \textit{Edge Attachment with Locality} (an inserted edge closing an open triangle): It is responsible for the edge destination.
    \item \textit{Node Lifetime} and \textit{Time Gap between Emitting Edges}: It is responsible for edge initiation.
    \item \textit{Node Arrival Rate}: It is responsible for the node arrival.
\end{itemize}
To model the individual behaviors, there are many candidate models for selection. For example, in edge attachment, the probability of a newcomer $u$ to connect the node $v$ can be proportional to $v$'s current degree or $v$'s current age or the combination. Based on fitting each model to the real-world observation under the supervision of MLE principle, the authors empirically choose the $\textit{random-random}$ model for edge attachment with locality, i.e., first, let node $u$ choose a neighbor $v$ uniformly and let $v$ uniform randomly choose $u$'s neighbor $w$ to close a triangle. And node lifetime and time gap between emitting edges are defined as follows.
\begin{equation}
    a(u) = t_{d(u)}(u) - t_{1}(u)
\end{equation}
where $a(u)$ stands for the age of node $u$, $t_{k}(u)$ is the time when node $u$ links its $k^{th}$ edge, $d_t(u)$ denote the degree of node $u$ at time $t$, and $d(u) = d_T(u)$. $T$ is the final timestamp of the data.
\begin{equation}
    \delta_{u}(d) = t_{d+1}(u) - t_{d}(u) 
\end{equation}
where $\delta_{u}(d)$ records the time gap between the current time and the time when that node emits its last edge.
Finding the node arrival is a regression process in~\cite{DBLP:conf/kdd/LeskovecBKT08}, for example, in Flickr graph $N(t) = exp(0.25t)$, and $N(t) = 3900t^{2} + 76000t -130000$ in LinkedIn graph.

In~\cite{DBLP:journals/corr/YangDC13,DBLP:conf/kdd/Park018}, the selection of edge attachment gets flourished where the authors propose several variants of edge attachment models for preserving the graph properties. With respect to the triangle closure phenomenon, several in-depth researches follow up. For example, in~\cite{DBLP:journals/tkdd/HuangDTYCF18}, researchers found that (1) \textit{the stronger the third tie (the interaction frequency of the closed edge) is, the less likely the first two ties are weakened}; (2) \textit{when the stronger the first two ties are, the more likely they are weakened}.

\subsection{High-Order Microscopic Laws} Hypergraph ego-network~\cite{comrie2021hypergraph} is a structure defined to model the high-order interactions involving an individual node. The star ego-network $T(u)$ is defined as follows.
\begin{equation}
    T(u) = \{s : (u \in s) \}, \forall s \in S
\end{equation}
where S is the set of all hyperedges (or simplices). Also, in ~\cite{comrie2021hypergraph}, there are other hypergraph ego-networks, like radial ego-network $R(u)$ and contracted ego-network $C(u)$. The relationship between them is as follows.
\begin{equation}
    T(u) \subseteq R(u) \subseteq C(u)
\end{equation}

In~\cite{comrie2021hypergraph}, authors observe that contiguous hyperedges (simplices) in an ego-network tend to have relatively large interactions with each other, which suggests that \textit{temporally adjacent high-order interactions have high similarity, i.e., the same nodes tend to appear in neighboring simplices}.

In~\cite{DBLP:conf/www/ZenoFN20}, authors try to model the temporal graph growth in terms of motif evolution activities. In brief, this paper investigates how many motifs change and what are the exact motif types in each time interval and fits the arrival rate parameter of each type of motif against the whole observed temporal graph.

\section{Law-Guided Research Tasks}
\label{sec:applications}
The discovered graph laws describe the graph property, which provides guidance to many downstreaming tasks. Some examples are discussed below.

\subsection{Graph Generation}
If not all, in most of graph law studies~\cite{DBLP:conf/kdd/LeskovecKF05,DBLP:journals/tkde/ZangCFZ18,DBLP:conf/kdd/DoYHS20,DBLP:conf/icdm/KookKS20,DBLP:conf/kdd/LeskovecBKT08,DBLP:conf/kdd/Park018,DBLP:conf/www/ZenoFN20}, after the law (i.e., evolution pattern) is discovered, a follow-up action is to propose the corresponding graph generative model to test whether there is a realizable graph generator could generate graphs while preserving the discovered law in terms of graph properties.
Also, graph generation tasks have impactful application scenarios like drug design and protein discovery~\cite{DBLP:journals/corr/abs-2307-08423}.

For example, in~\cite{DBLP:conf/kdd/LeskovecKF05}, the Forest Fire model is proposed to preserve the macroscopic graph law while larges preserve the discovered evolution pattern.
\begin{itemize}[leftmargin=*]
    \item First, node $v$ first chooses an ambassador (i.e., node $w$) uniformly random, and establish a link to $w$;
    \item Second, node $v$ generates a random value $x$, and selects $x$ links of node $w$, where selecting in-links $r$ times less than out-links;
    \item Third, node $v$ forms links to $w$'s neighbors; this step executes recursively (neighbors of neighbors) until $v$ dies out.
\end{itemize}
This proposed Forest Fire model holds the following graph properties most of time.
\begin{itemize}[leftmargin=*]
    \item \textit{Heavy-tailed In-degrees}: The highly linked nodes can easily get reached, i.e., “rich get richer”.
    \item \textit{Communities}: A newcomer copies neighbors of its ambassador.
    \item \textit{Heavy-tailed Out-degrees}: The recursive nature produces large out-degree.
    \item \textit{Densification Law}: A newcomer will have a lot of links near the community of its ambassador.
    \item \textit{Shrinking Diameter}: It may not always hold.
\end{itemize}

In~\cite{DBLP:conf/kdd/LeskovecBKT08}, the authors combine the microscopic edge destination model, edge initiation model, and node arrival rate together, to model the real-world temporal network's growth. The parameters of these three models are fitted against the partial observation. i.e., $G_{\frac{T}{2}}$, which is the half of the entire evolving graph. Then they three produce the residual part of $G_{T}'$. Finally, the generated $G_{T}'$ is compared with the ground truth $G_{T}$, to see if the growth pattern is fully or near fully captured by these microscopic models. The procedures are stated as follows.

\begin{itemize}[leftmargin=*]
    \item First, nodes arrive using the node arrival function obtained from $G_{\frac{T}{2}}$;
    \item Second, node $u$ arrives and samples its lifetime $a$ from the age distribution of $G_{\frac{T}{2}}$;
    \item Third, node $u$ adds the first edge to node $v$ with probability proportional to node $v$'s degree;
    \item Fourth, node $u$ with degree $d$ samples a time gap $\delta$ from the distribution of time gap in $G_{\frac{T}{2}}$;
    \item When a node wakes up, if its lifetime has not expired yet, it creates a two-hop edge using the "random-random" triangle closing model;
    \item If a node’s lifetime has expired, then it stops adding edges; otherwise, it repeats from Step 4.
\end{itemize}

The generated graph $G_{T}'$ is tested based on the comparison with the ground truth $G_{T}$, in terms of degree distribution, clustering coefficient, and diameter distribution. Taking the Flickr graph for example, the generated graph is very similar to the ground truth with aforementioned metrics~\cite{DBLP:conf/kdd/LeskovecBKT08}.

\subsection{Link Prediction}
To learn node representation vectors for predicting links between node pairs and contributing latent applications like recommender systems, CAW-N~\cite{DBLP:conf/iclr/WangCLL021} is proposed by inserting causal anonymous walks (CAWs) into the representation learning process. The CAW is a sequence of time -aware adjacent nodes, the authors claim that the extracted CAW sequence obeys the triadic closure law. To be specific, the temporal opening and closed triangles can be preserved in the extracted CAW sequence $W$. Further, to realize the inductive link prediction, CAW-N replaces the identification of each node in $W$ with the relative position information, such that the CAW sequence $W$ is transferred into anonymous $\hat{W}$.
Then, the entire $\hat{W}$ is inserted into an RNN-like model and gets the embedding vector of each node, the loss function states as follows.
\begin{equation}
    enc(\hat{W}) = \textbf{RNN}(\{f_{1}(I_{CAW}(w_{i})) \oplus f_{2}(t_{i-1} - t_{i})\}_{i=0,1,\ldots, |\hat{W}|})
\end{equation}
where $I_{CAW}(w_{i})$ is the anonymous identification of node $i$ in $\hat{W}$, $f_{1}$ is the node embedding function realized by a multi-layer perceptron, $f_{2}$ is the time kernel function for representing a discrete time by a vector, and $\otimes$ denotes the concatenation operation. The training loss comes from predicting negative (disconnected) node pairs and positive (connected) node pairs.


Also, there are some related link prediction models based on the guidance of static graph laws during the representation learning process, for example, SEAL~\cite{DBLP:conf/nips/ZhangC18} and HHNE~\cite{DBLP:conf/aaai/WangZS19a}.

In the SEAL framework~\cite{DBLP:conf/nips/ZhangC18}, for each target link, SEAL extracts a local enclosing subgraph around it, and uses a GNN to learn general graph structure features for link prediction. The corresponding graph parameters include but are not limited to 
\begin{itemize}[leftmargin=*]
    \item \textit{Common Neighbors}: Number of common neighbors of two nodes.
    \item \textit{Jaccard}: Jaccard similarity on the set of neighbors of two nodes.
    \item \textit{Preferential Attachment}: The product of the cardinal of the sets of neighbors of two nodes.
    \item \textit{Katz Index}: The summarization over the collection of paths of two nodes.
\end{itemize}

\section{Some New Observation Space and Newly Discovered Graph Parameters}
\label{sec: new params}

\subsection{New Different Spaces}
In~\cite{DBLP:conf/kdd/EikmeierG17}, the power law is revisited based on the eigendecomposition and singular value decomposition to provide guidance on the presence of power laws in terms of the degree distribution, singular value (of adjacency matrix) distribution, and the eigenvalue (of Laplacian matrix) distribution. The authors~\cite{DBLP:conf/kdd/EikmeierG17} discovered that (1) degree distribution, singular value distribution, and eigenvalue distribution follow power law distribution in many real-world networks they collected; (2) and a significant power law distribution of degrees usually indicates power law distributed singular values and power law distributed eigenvalues with a high probability.

\subsection{New Parameters}
Currently, if not all, most graph law research focuses on the traditional graph properties, like the number of nodes, number of edges, degrees, diameters, eigenvalues, and singular values. Here, we provide some recently proposed graph properties, although they have not yet been tested on the scale for fitting the graph law on real-world networks.

The local closure coefficient~\cite{DBLP:conf/wsdm/YinBL19} is defined as the fraction of length-2 paths (wedges) emanating from the head node (of the wedge) that induce a triangle, i.e., starting from a seed node of a wedge, how many wedges are closed. According to~\cite{DBLP:conf/wsdm/YinBL19}, features extracted within the constraints of the local closure coefficient can improve the link prediction accuracy. The local closure efficient of node $u$ is defined as follows.
\begin{equation*}
    H(u) = \frac{2T(u)}{W^{h}(u)}
\end{equation*}
where $W^{(h)}(u)$ is the number of wedges where $u$ stands for the head of the wedge, and $T(u)$ denotes the number of triangles that contain node $u$.

The density of states (or spectral density)~\cite{DBLP:conf/kdd/DongBB19} is defined as follows.
\begin{equation}
    \mu(\lambda) = \frac{1}{N} \sum ^{N}_{i=1}\delta(\lambda - \lambda_{i}),~ \int f(\lambda)\mu(\lambda) = \text{trace}(f(H))
\end{equation}
where $H$ denotes any symmetric graph matrix, $\lambda_{1}, \ldots, \lambda_{N}$ denote the eigenvalues of $H$ in the ascending order, $\delta$ stands for the Dirac delta function and $f$ is any analytic test function.

\section{Related Work}
\label{sec:related_work}
To the best of our knowledge, there are only a few survey papers on graph laws, with none published after 2022, marking the beginning of the foundation model era. A 2006 survey~\cite{DBLP:journals/csur/ChakrabartiF06} primarily focused on graph laws for mining patterns, discussing the Densification Law and Shrinking Law.
In 2016, another survey~\cite{drakopoulosa2016large} shifted its focus towards the generation of large graphs using different graph modeling methods, including the Erdos-Renyi model, Watts-Strogatz model, and Albert-Barabasi model.
More recently, in 2019, the authors in~\cite{DBLP:journals/csur/DrobyshevskiyT20} offered a broader perspective on random graph modeling, covering generative, feature-driven, and domain-specific approaches.
In contrast to these earlier surveys, which were published before the advent of graph neural networks and prior to the discovery of several significant graph laws~\cite{DBLP:conf/wsdm/YinBL19,DBLP:conf/kdd/DongBB19,DBLP:conf/www/ZenoFN20,DBLP:conf/kdd/DoYHS20,DBLP:conf/icdm/KookKS20,comrie2021hypergraph,DBLP:conf/aaai/WangZS19a}, our work represents the first survey to explore the potential of graph laws in the context of foundation models. We emphasize how graph laws can address domain inconsistencies across different graph data types and contribute to multimodal representation learning. Additionally, this survey is the first to offer an overview of high-order graph laws and heterogeneous graph laws, marking a novel contribution to the literature.

\section{Future Directions}
\label{sec:future_directions}
Here, we list several interesting research directions of graph parametric representation in modern graph research.

\subsection{Graph Laws on Temporal Graphs}
Discovering accurate temporal graph laws from real-world networks heavily relies on the number of networks and the size of networks (e.g., number of nodes, number of edges, and time duration). However, some of the temporal graph law studies mentioned above usually consider the number of graphs ranging from 10 to 20, when they discover the evolution pattern. The existence of time-dependent structure and feature information increases the difficulty of collecting real-world temporal graph data. To obtain robust and accurate (temporal) graph laws, we may need a considerably large amount of (temporal) network data available. Luckily, we have seen some pioneering work like TGB~\cite{DBLP:journals/corr/abs-2307-01026} and TUDataset~\cite{Morris+2020}.

\subsection{Graph Laws on Heterogeneous Networks}
Though many graph laws have been proposed and verified on homogeneous graphs, real-world networks are usually heterogeneous \cite{DBLP:journals/tkde/ShiLZSY17} and contain a large number of interacting, multi-typed components. While the existing work \cite{DBLP:conf/aaai/WangZS19a} only studied 2 datasets to propose and verify the heterogeneous graph power law, the potential exists for a transition in graph laws from homogeneous networks to heterogeneous networks, suggesting the presence of additional parameters contributing to the comprehensive information within heterogeneous networks. For example, in an academic network, the paper citation subgraph and the author collaboration subgraph may have their own subgraph laws which affect other subgraphs' laws. Furthermore, Knowledge graphs, as a special group of heterogeneous networks, have not yet attracted much attention from the research community to study their laws.

\subsection{Transferability of Graph Laws}
As we can see in the front part of the paper, many nascent graph laws are described verbally without the exact mathematical expression, which hinders the transfer from the graph law to the numerical constraints for the representation learning process. One latent reason for this phenomenon is that selecting appropriate models and parameters and fitting the exact values of parameters on large evolving graphs are very computationally demanding.

\subsection{Taxonomy of Graph Laws}
After we discovered many graph laws, is there any taxonomy or hierarchy of those? For example, graph law A stands in the superclass of graph law B, and when we preserve graph law A during the representation, we actually have already preserved graph law B. For example, there is a hierarchy of different computer vision tasks, recently discovered~\cite{DBLP:conf/cvpr/ZamirSSGMS18}. And corresponding research on graph law development seems like a promising direction.



\subsection{Domain-Specific Graph Laws}
Since graphs serve as general data representations with extreme diversity, it is challenging to find universal graph laws that fit all graph domains because each domain may be internally different from another \cite{DBLP:journals/corr/abs-2308-14522}. In fact, in many cases, we have prior knowledge about the domain of a graph, which can be a social network, a protein network, or a transportation network. Thus, it is possible to study the domain-specific graph laws that work well on only a portion of graphs and then apply the graph laws only on those graphs.

\subsection{Graph Laws with LLMs}
\label{sec:graph_llm}
In the background of large language models (LLMs) developments, an interesting question attracts much research interest nowadays, i.e., \textbf{can LLMs replace GNNs as the backbone model for graphs?} To answer this question, many recent works show the great efforts~\cite{DBLP:journals/corr/abs-2309-16595,DBLP:conf/eacl/YeZWXZ24,DBLP:conf/iclr/HeB0PLH24}, where the key point is how to represent the structural information as the input for LLMs.

For example, Instruct-GLM~\cite{DBLP:conf/eacl/YeZWXZ24} follows the manner of instruction tuning and makes the template $\mathcal{T}$ of a 2-hop connection for a \textit{central node} $v$ as follows.
\begin{equation}
    \mathcal{T}(v, \mathcal{A}) = \{v\} \text{~is connected with~} \{|v_{2}|_{v_{2} \in \mathcal{A}^{v}_{2}}\} \text{~within two hops.~}
\end{equation}
where $\mathcal{A}^{v}_{k}$ represents the list of node $v$'s $k$-hop neighbors.

As discussed above, the topological information (e.g., 1-hop or 2-hop connections) can serve as external modality information to contribute to (e.g., through prompting) the reasoning ability of large language models (LLMs)~\cite{DBLP:journals/corr/abs-2309-16595} and achieve state-of-the-art on low-order tasks like node classification and link prediction.

\section{Conclusion}
Motivated by LLM's need to understand graphs, we first review the concepts and development progress of graph parametric representations (i.e., graph laws) from different perspectives like macroscope and microscope, low-order and high-order connections, and static and temporal graphs. We then discuss various real-world application tasks that can benefit the study of graph parametric representations. Finally, we envision the latent challenges and opportunities of graph parametric representations in modern graph research with several interesting and possible future directions.




\newpage

\bibliographystyle{named}
\bibliography{reference}

\end{document}